\documentclass[11pt]{article}
\usepackage{geometry}
\usepackage{coling2020}
\usepackage{times}
\usepackage{url}
\usepackage{latexsym}
\usepackage{microtype}
\usepackage{graphicx}
\usepackage{svg}
\usepackage{amsmath}

\usepackage{tabularx}
\colingfinalcopy 
\newcounter{example}[section]

\title{NoPropaganda at SemEval-2020 Task 11: A Borrowed Approach to Sequence Tagging and Text Classification}

\author{Ilya Dimov \\
  ABBYY ARD NLP\\
  Lomonosov Moscow State University\\
  {\tt ilia.dimov@abbyy.com} \\\And
  Vladislav Korzun \\
  ABBYY ARD NLP\\
  MIPT ABBYY Lab\\
  {\tt vladislav.korzun@abbyy.com} \\\AND
  Ivan Smurov \\
  ABBYY ARD NLP \\
  MIPT ABBYY Lab\\
  {\tt ivan.smurov@abbyy.com} \\
  }

\date{}

\begin{document}
\maketitle

\medskip
\medskip
\medskip
\medskip
\medskip
\begin{abstract}
  This paper describes our contribution to SemEval-2020 Task 11: \textit{Detection Of Propaganda Techniques In News Articles}. We start with simple LSTM baselines and move to an autoregressive transformer decoder to predict long continuous propaganda spans for the first subtask. We also adopt an approach from relation extraction by enveloping spans mentioned above with special tokens for the second subtask of propaganda technique classification. Our models report an F-score of 44.6\% and a micro-averaged F-score of 58.2\% for those tasks accordingly.
\end{abstract}

\section{Introduction}
\label{intro}
In recent years natural language processing has experienced rapid development. In particular, application of deep learning to NLP \cite{collobert2011natural}, the introduction of pre-trained context-independent word embeddings such as word2vec \cite{mikolov2013a,mikolov2013b}, GloVe \cite{pennington2014glove} and fastText \cite{fasttext}, and utilization of RNN-based \cite{hochreiter1997long} pipelines such as CharCNN-BLSTM-CRF \cite{lample2016neural,ma2016endtoend} allowed to improve SOTA for an overwhelming majority of NLP tasks. In 2018 a monumental breakthrough happened when context-dependent embeddings based on pre-trained language models emerged, such as ELMo \cite{peters2018deep}, ULMFit \cite{ulmfit}, and BERT \cite{DBLP:journals/corr/abs-1810-04805}. The resulting boost in the models' performance cannot be underestimated, amounting to a 30\% relative error reduction rate.

Researchers compare this latest development to the introduction of ImageNet to Computer Vision. Furthermore, as breakthroughs in CV allowed such technologies as DeepFake to arise, similar processes are happening to NLP: the authors of GPT2 \cite{radford2019language} anticipated that their model could be used for malicious purposes such as fake news generation or even impersonation. Given that at present, the problem of media swaying public opinion to the benefit of certain parties is more acute than ever, we must keep our news fair and unbiased; thus,  techniques for fake news and propaganda detection need to be developed. SemEval-2020 task 11 is an essential step in that direction.

Another noteworthy consequence of contextualized embeddings introduction is the generalization of models. While many models are developed for a particular task, they can be effectively utilized in order to solve other tasks. Our main contribution is the utilization of two such models for new tasks, namely LaserTagger \cite{malmi2019encode}, initially developed for summarization, which is used for span identification, while R-BERT \cite{wu2019enriching}, originally developed for relation extraction, is used for span classification. 

Section 2 contains the description of our systems\footnote{The code is released at \url{https://github.com/hawkeoni/Semeval2020_task11.}}, section 3 contains evaluation results and error analysis, while part 4 contains the concluding remark. 
\blfootnote{
    %
    %
    \hspace{-0.65cm}  
    This work is licensed under a Creative Commons Attribution 4.0 International Licence. Licence details: \url{http://creativecommons.org/licenses/by/4.0/}.
}
\section{System description}
All of the following models based on \textbf{BERT} \cite{DBLP:journals/corr/abs-1810-04805} use \textbf{BERT\textsubscript{BASE}} pretrained model weights. Hyperparameter tuning and cross-validation was performed on 5 folds split by articles.
The data used for training and evaluation consists of various news articles and is described in detail in \cite{EMNLP19DaSanMartino}. 
The tasks will be briefly explained in the following sections; their full description can be found in the \cite{DaSanMartinoSemeval20task11}.

\subsection{Span Identification}
The goal of this task is to find continuous spans of text which contain any propaganda technique. Consider the following example, which has \textit{loaded language} marked up in bold and double squared brackets:

\begin{quote}
It’s essentially an admission of guilt, given that it is  \textbf{[[absolutely ludicrous]]} to think that “national security” would be threatened by the release of the CIA’s long-secret JFK-assassination-related records.
\end{quote}

We treat this task as a binary sequence tagging where each token is assigned a label: 0 - for normal text, 1 - for propaganda text. Although there are spans which overlap multiple sentences, we decided to split such spans and solve the problem on a sentence level. This approach has a distinct disadvantage, as the connection between sentences is lost.

At first we started with several baseline bidirectional LSTM models with linear layer for token class prediction: 
\begin{itemize}
    \item \textbf{biLSTM\textsubscript{GLOVE + charLSTM}} - a bidirectional LSTM over GloVe embeddings and a character-level LSTM. This is a simple sequence tagging model proposed originally in \cite{lample2016neural}.
    \item \textbf{biLSTM\textsubscript{ELMO}} - a bidirectional LSTM over ELMo embeddings from \cite{peters2018deep}.
\end{itemize}

Preliminary results of those models achieved an F-score of 24-26\% on the development set. Postprocessing model predictions by labeling all tokens between the first sentence-wise positive label and the last sentence-wise positive label as propaganda  pushed F-score to 34.5\%. A CRF instead of a linear layer provided no significant benefit.

Our next model BERT\textsubscript{LINEAR}, which consists of a linear layer over \textbf{BERT\textsubscript{BASE}}, achieved a stronger baseline with F-score of 40.8\% without any postprocessing. However, unlike in previous models postprocessing only made things worse, degrading prediction quality. 

Generating token tags with a linear layer has one big flaw: predictions for each token are made independently. CRF mitigates this problem and improves the metrics, but we decided to employ \textbf{LaserTagger} - an autoregressive transformer decoder from \cite{malmi2019encode}. One of the critical ideas of \textbf{LaserTagger} is directly consuming corresponding encoder activation without learning encoder-decoder attention weights. \textbf{LaserTagger} achieved an even stronger baseline with an F-score of 42\% on development set, which at the time was only several points away from top-performing teams. It was decided to keep this model and extensively tune it during further experiments.

\begin{figure}[h!]
    \includegraphics[scale=0.8]{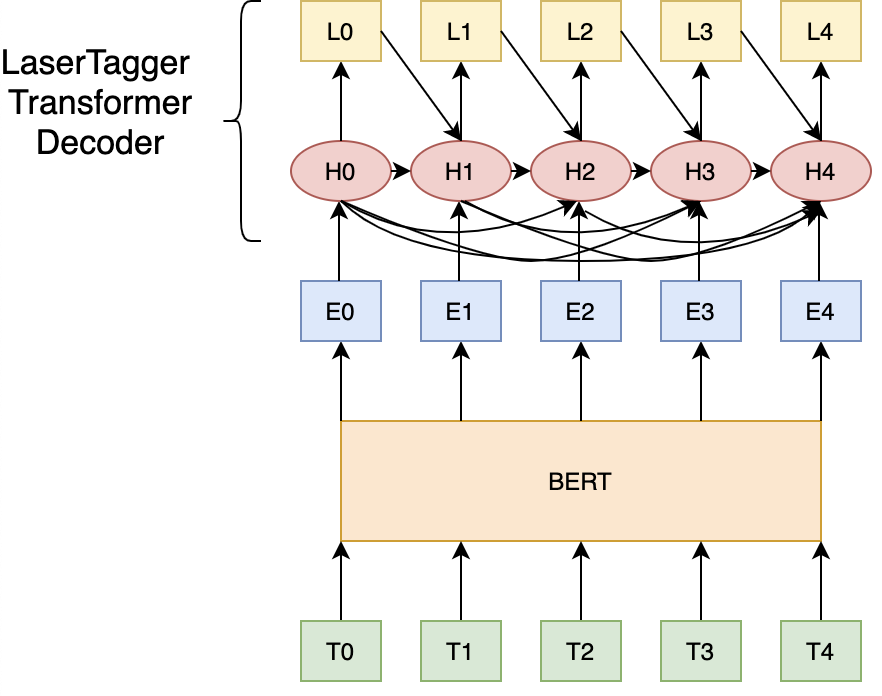}
    \caption{LaserTagger model architecture. T\textsubscript{i} are token embeddings for \textbf{BERT}, E\textsubscript{i} are \textbf{BERT} outputs, H\textsubscript{i} represent LaserTagger hidden states and L\textsubscript{i} are predicted labels.}
\end{figure}

We have employed several techniques to boost model performance:
\begin{itemize}
    \item We fine-tune \textbf{BERT} as a masked language model on task corpus for 3 epochs without next sentence prediction loss. This provides no noticeable increase in F-score.
    \item Teacher forcing (scheduled sampling) \cite{bengio2015scheduled} is a method of improving performance for autoregressive models. As \textbf{LaserTagger} predictions explicitly depend on predictions from the previous step, errors accumulated earlier may lead to a completely wrong result. Teacher forcing aims to close the gap between training and inference by supplying a model with its own (possibly wrong) predictions at a specific rate during training. This allows the model to explore more of its state space during training, which in turn increases its robustness during inference. We linearly decay teacher forcing rate from 1 (only correct labels) to 0 (only model predictions) during training.
    The downside of this approach is a significant slowdown in training as we do not utilize the transformer's capability to process the whole sequence at once.
    \item Label smoothing was used to lower the number of false-positive predictions and mitigating minor data discrepancies like inconsistent inclusion of dots and quotations in propaganda spans, which were caused by limiting spans to sentences and other preprocessing imperfections.
\end{itemize}

Our final submission was a single \textbf{LaserTagger} model with \textbf{BERT\textsubscript{BASE}} encoder, single-layered decoder with hidden dimension of 128 and 4 attention heads. The model was trained on an effective batch size of 32 achieved by gradient accumulation over every 2 steps; we used Adam optimizer with learning rate of 2e-5 and learning rate warmup for 10\% of the steps. Teacher forcing was set to linearly decay from 1 to 0 for the duration of the whole training. This model achieved F-score of 46.1\% on development set and 44.6\% on test set.

\subsection{Technique Classification}
The second subtask was to classify spans from previous subtasks into 14 classes of propaganda techniques. As previously mentioned, spans can overlap several sentences, so for this subtask, we had to use all overlapping sentences for correct classification. As spans with different techniques may overlap, or a single span may be associated with several techniques, we treat this subtask as a multilabel classification problem.

Our main idea was taken from \textbf{R-BERT} \cite{wu2019enriching}, where special tokens were inserted around entities to highlight them in the task of relation extraction. As in \textbf{R-BERT}, we surround propaganda spans with special tokens, namely the \textbf{\^} token. \textbf{BERT} representations of propaganda span tokens are then averaged or taken as a weighted sum based on results of the linear layer applied to \textbf{BERT} token activations in the following way:
let $e_i...e_{i+k}\in R^n$ be \textbf{BERT} representations of propaganda tokens in the current span and $w \in R^{n}, b \in R$ - learnable parameters. Then the weighted sum is calculated as $\sum_{j=i}^{i+k} \alpha_j e_j$, where $\alpha_j$ is the result of softmax function over scalar product of $e_j$ and $w$: 

\begin{equation}
\alpha_j = \frac{\exp{((w, e_j) + b)}}{\sum_{j=i}^{i+k}{\exp{((w, e_j) + b)}}}
\end{equation}

The resulting vector is concatenated with \textbf{BERT} output of the \textbf{[CLS]} token for final multilabel prediction. We also combine these approaches with previously mentioned \textbf{BERT} intask finetuning \cite{sun2019fine}, which proves to be fruitful for this task as it is not too dissimilar to text classification. We additionally train a simple \textbf{BERT} model, which predicts technique class using only \textbf{[CLS]} token representation. All models were trained with linear warmup for 10\% of the training steps using the Adam optimizer with learning rate of 2e-5.

Our final submission for task TC consists of predictions averaged from several models, which are described in Section 3.2.

\begin{figure}[h!]
    \centering
    \includegraphics[scale=0.75]{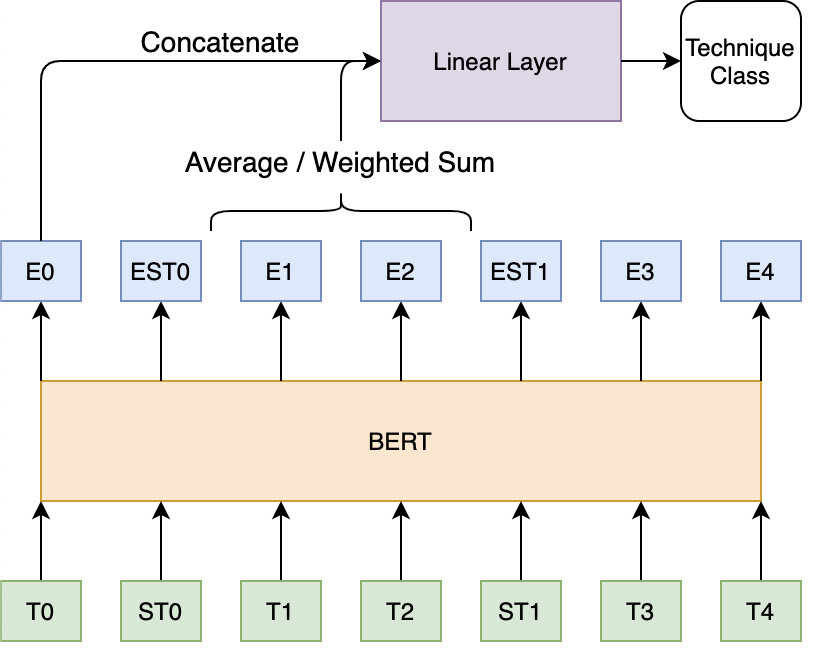}
    \caption{Task TC model architecture based on R-BERT. T\textsubscript{i} are token embeddings for \textbf{BERT}; ST\textsubscript{0} and ST\textsubscript{1} represent special tokens used to highlight the propaganda span; E\textsubscript{i} and EST\textsubscript{i} are \textbf{BERT} outputs}.
\end{figure}

\section{Experimental Results}
\subsection{Span Identification Results}
\begin{table}[h]
\begin{center}
\begin{tabular}{|l|rrr|}
\hline \bf Model & \bf F1 & \bf Precision & \bf Recall \\ \hline
biLSTM\textsubscript{GLOVE + charLSTM} & 0.249 & 0.302 & 0.212 \\
biLSTM\textsubscript{GLOVE + charLSTM} + postprocessing & 0.349 & 0.304 & 0.411 \\
biLSTM\textsubscript{ELMO} + postprocessing &  0.345 & 0.327 & 0.367 \\
BERT\textsubscript{LINEAR} & 0.408 & 0.357 & 0.477 \\
BERT + CRF & 0.362 & 0.301 & 0.454 \\
BERT + LSTM + CRF & 0.414 & 0.363	& 0.482 \\
LaserTagger & 0.420 & 0.384 & 0.464 \\
LaserTagger + TF & 0.451 & \textbf{0.423} & 0.482 \\
LaserTagger + TF + LS & \textbf{0.461} & 0.406 & \textbf{0.533} \\
\hline
LaserTagger + TF + LS (test)& 0.446	& 0.556 & 0.373 \\
\hline
\end{tabular}
\end{center}
\caption{\label{font-table} Task SI model results on development and test sets. TF and LS stand for teacher forcing and label smoothing  accordingly.}
\end{table}

As Table 1 shows \textbf{BERT}-based approaches significantly outperform suggested LSTM baselines. Main improvements over simple \textbf{BERT\textsubscript{LINEAR}} model is \textbf{LaserTagger} decoder, which eliminates the flaw of independent tag prediction. It was further improved by linearly decreasing teacher forcing rate over the training and label smoothing. Our model scored 7th place out of 35 in the final table of the competition.

\begin{table}
\begin{tabularx}{\textwidth} {X}

\hline 
Boundary errors  \\ 
\hline
\textit{Gold:} This is a \textbf{[[smear]]}, plain and simple. \\
\textit{Prediction:} This is \textbf{[[a smear]]}, plain and simple. \\ 
\textit{Gold:} Luke Harding and the Guardian Publish Still More \textbf{[[Blatant MI6 Lies]]} \\
\textit{Prediction:} Luke Harding and the Guardian Publish Still More \textbf{[[Blatant]]} MI6 Lies \\
\hline
Quotation and punctuation errors \\
\hline
\textit{Gold:} While trying to appear as a moderate, "Gillum is a Progressive" and "He is a part of the \textbf{[[crazy, crazy, crazies]]}." \\
\textit{Prediction:} While trying to appear as a moderate, "Gillum is a Progressive" and "\textbf{[[He is a part of the crazy, crazy, crazies]]}." \\
\\
\textit{Gold:} "\textbf{[[I don’t care who you are]]}, determine the scope and the parameters of an investigation. \\
\textit{Prediction:} "\textbf{[[I don’t care who you are, determine the scope and the parameters of an investigation]]}. \\
\\
\textit{Gold:} “\textbf{[[This is devastating]]} for our Democratic Latino community. \\
\textit{Prediction:} “\textbf{[[This is devastating for our Democratic Latino community.]]} \\

\hline
\caption{\label{font-table} LaserTagger + TF + LS errors.}
\end{tabularx}
\end{table}

Table 2 illustrates the errors of our final model. Besides sometimes incorrectly adding or losing boundary propaganda span tokens, \textbf{LaserTagger + TF + LS} seems to have a problem with quotation and punctuation. When it encounters quoted text or a part of a sentence separated by commas, it either labels the whole highlighted area as propaganda or misses it completely. This may be partly attributed to the nature of quotes - speech often conveys more emotional words than written text. Another problem is that when the model finds comma, it often closes the propaganda span and does not reopen it.

\subsection{Technique Classification Results}
\begin{table}[h]
\begin{center}
\begin{tabular}{|l|r|}
\hline \bf Model & \bf micro-averaged F-measure \\ \hline
BERT\textsubscript{CLS} & 0.573 \\
R-BERT & 0.592 \\
R-BERT\textsubscript{FT} & 0.584 \\
R-BERT\textsubscript{W} & 0.590 \\
R-BERT\textsubscript{W + FT} & 0.590 \\
All models & \textbf{0.606} \\
\hline
All models (test) & 0.582 \\
\hline
\end{tabular}
\end{center}
\caption{\label{font-table} Task TI model results on development and test sets. W and FT stand for weighted sum instead of averaging \textbf{BERT} token representation and intask finetuning accordingly.}
\end{table}

All models from the table use the \textbf{\^} special symbol to highlight propaganda spans because more than 20\% of the sentences contain multiple propaganda spans. For example, if the model has 5 tokens $e_1...e_5$ and two propaganda spans $e_1, e_2$ and $e_4, e_5$ we would create two training samples with the use token \textbf{\^}, which we will write for readability as \textbf{st}:
$st, e_1, e_2, st, e_3, e_4, e_5$ and $e_1, e_2, e_3, st, e_4, e_5, st.$
This allows us to point our model to a specific propaganda span inside a sentence.

Models from Table 3 are the following:
\begin{itemize}
    \item \textbf{BERT\textsubscript{CLS}} - a model that predicts technique solely from \textbf{[CLS]} token embedding.
    \item \textbf{R-BERT} - a model that predicts technique from  \textbf{[CLS]} token embedding and averaged propaganda span representation;
    \item \textbf{R-BERT\textsubscript{W}} - a model that predicts technique from  \textbf{[CLS]} token embedding and weighted propaganda span representation;
\end{itemize}

It is important to understand that due to class imbalance models with the same micro-averaged F-score correspond to different F-scores over specific techniques. While ensembling the models increases overall quality, two classes, namely \textbf{"Bandwagon, Reduction Ad Hitlerum"} and \textbf{"Whataboutism, Straw Men, Red Herring"}, are entirely missed by all our models. This is not surprising: those techniques have a very limited amount of positive samples, which is the reason they were merged into classes. Class wise F-score seems to respond to the quantity of corresponding class in training dataset: our model performs best on \textbf{Loaded Language} and \textbf{Name Calling, Labeling} followed by \textbf{Flag-Waving} and \textbf{Doubt}.

We did not implement anything to boost performance on underrepresented classes, while some of the other teams found a way to get an F-score of 20\% and higher. Our model scored 6th place out of 31 in the final table of the competition.

\subsection{Dataset Exploration}
During the competition, most teams had a substantial shift between recall and precision: precision was 10-20\% lower than recall. On the contrary, test dataset evaluation resulted in a different picture: top-performing teams' precision is 5-15\% higher than recall. Additionally, on task TC all teams suffered a downgrade in micro-averaged F-score of approximately 5\%.

This can be attributed to the fact that the datasets have imbalanced topic distributions. While the training dataset is not very large, it is well-rounded and covers a wide range of topics. The development set is focused on the US news, while the test set seems to have almost an even split between the UK and the US news.

\section{Conclusion}
In this work, we propose to use two models: autoregressive transformer decoder for sequence tagging and \textbf{R-BERT}-like approach for propaganda spans technique classification. The former is taken from such tasks as sentence fusion, sentence splitting, and abstractive summarization, while the latter was originally used for relation extraction.

Instead of exploring various pretrained transformers like \textbf{BERT\textsubscript{LARGE}}, \textbf{RoBERTa}, \textbf{ALBERT} and \textbf{XLNet} or engineering domain-specific features and approaches, we focused on improving architectures and trying to push them as far as possible.

In the future the results of this task may be used for propaganda detection in various media as well as unbiasing texts and making them fair.


\bibliographystyle{coling}
\bibliography{semeval2020}

\end{document}